\documentclass{article}
\usepackage{spconf,amsmath,graphicx,hyperref}
\usepackage{amssymb, booktabs}
\usepackage{multirow}
\usepackage{xcolor}

\title{Neural Surrogate Modeling of Ultrasonic Wave Propagation in Heterogeneous Media}
\name{
Burla Nur Korkmaz$^{1}$*,
Petro Moilanen$^{2}$,
Ari Salmi$^{2}$,
Harsha Vardhan Tetali$^{1}$,
Arto Klami$^{1}$%
\thanks{* Corresponding author: burla.korkmaz@helsinki.fi}
}

\address{
$^{1}$ Department of Computer Science, University of Helsinki, Finland\\
$^{2}$ Department of Physics, University of Helsinki, Finland
}
\begin{document}
%
\maketitle
\begin{abstract}

Neural surrogates provide computationally efficient auto-regressive modeling of wave propagation, enabling e.g. fast inverse solutions. Despite recent progress, the current surrogates are not yet sufficient for the needs of ultrasonic sensing, characterized by high frequency and heterogeneous propagation media. We show how methods building on Fourier Neural Operators (FNO) fail to capture essential signal features, such as reflections, whereas U-Net -based methods learn the wave fields well but lack the ability to transfer information across different spatial scales. We then propose a new surrogate that captures the best of the two worlds, by integrating FNOs and local-integral resampling into a U-Net architecture. We show it can be trained on limited data, and that the predictions are robust to input noise and work across spatial scales. 
\end{abstract}
\section{Introduction}
\label{sec:intro}

Ultrasound is ubiquitously used for non-destructive imaging of
the internals of structures in structural health monitoring (SHM)~\cite{rose2004ultrasonic, gharib2025self, korkmaz2026physical} and medical imaging~\cite{khan2022unfolding, xiao2025beamforming, ruan2024delineation, yan2025untrained}.
In these tasks, the key challenge is modeling and understanding how high-frequency acoustic waves propagate in highly heterogeneous media, with abundant reflections and refractions from material interfaces and external boundaries. For a known structure, the wavefield can be simulated numerically using, e.g.,finite-element method (FEM) or finite-difference method (FD), but this requires excessive memory and computation (hours, days or even weeks)
for time-domain solvers of realistic time-scales \cite{moilanen2024software}. Moreover, a single simulation is not enough; we typically need repeated simulations for alternative structures within an inverse algorithm \cite{guasch2020full, zhang2024cross}.

Neural surrogates, flexible machine learning models trained to approximate wave propagation, offer a highly promising alternative by amortizing the cost \cite{sharma2022deep}. Once trained, these models can simulate the wave propagation in new structures in a fraction of the simulation time, often in the scale of seconds, making broad range of inverse problems practical, including gradient-based solvers that require differentiable integrator \cite{augenstein2023neural}. However, this hinges on the neural surrogates being sufficiently accurate.  
Although they have demonstrated impressive performance in many scientific domains \cite{kovachki2023neural}, there are still open challenges in approximating ultrasound propagation in heterogeneous media due to the high frequency and the high numerical accuracy required for the precise boundary effects at the interfacing regions.
To address this, we need neural surrogates that better model wave propagation in heterogeneous media.

Prior methods 
rely on two main surrogate families, Fourier Neural Operators (FNO)~\cite{li2021fourier} and U-Nets~\cite{cciccek20163d, ronneberger2015u}. Neural operators, like FNOs, model directly the underlying continuous system and are hence invariant to the discretization resolution, but for now the success of FNO-based methods in acoustic wavefield simulations has been limited \cite{huang2025learned,zhang2023learning,li2023solving},
likely due to jump discontinuities at the material interfaces due to the Gibbs phenomenon \cite{gottlieb1997gibbs}.
U-Net based methods \cite{li2025steady,stender2023u} that construct progressively coarser representation to separate modeling of local and global phenomena address this aspect better, but they lack the operator framing and only learn the discretized system at one resolution.
Recently,~\cite{davis2026fourier} theoretically showed the advantage of introducing residual connections in FNO for modeling functions with jump discontinuities: this makes case for a hybrid architecture of Fourier layers within a U-Net structure. 
Already prior to this, U-NO~\cite{rahman2023uno} integrated FNOs into a U-Net architecture, whereas
CNO~\cite{raonic2023convolutional} considered convolutional neural operators with similar architecture.
We build on this line of research and  \emph{craft a combination of U-Net and FNO with modified resolution invariant shrinking and expanding operations adapted from~\cite{liu_schiaffini2024neural}}, shown to outperform the only
previous combination of U-Net and FNO~\cite{rahman2023uno} both quantitatively and qualitatively.

\begin{figure}
    \centering
    \includegraphics[width=\linewidth]{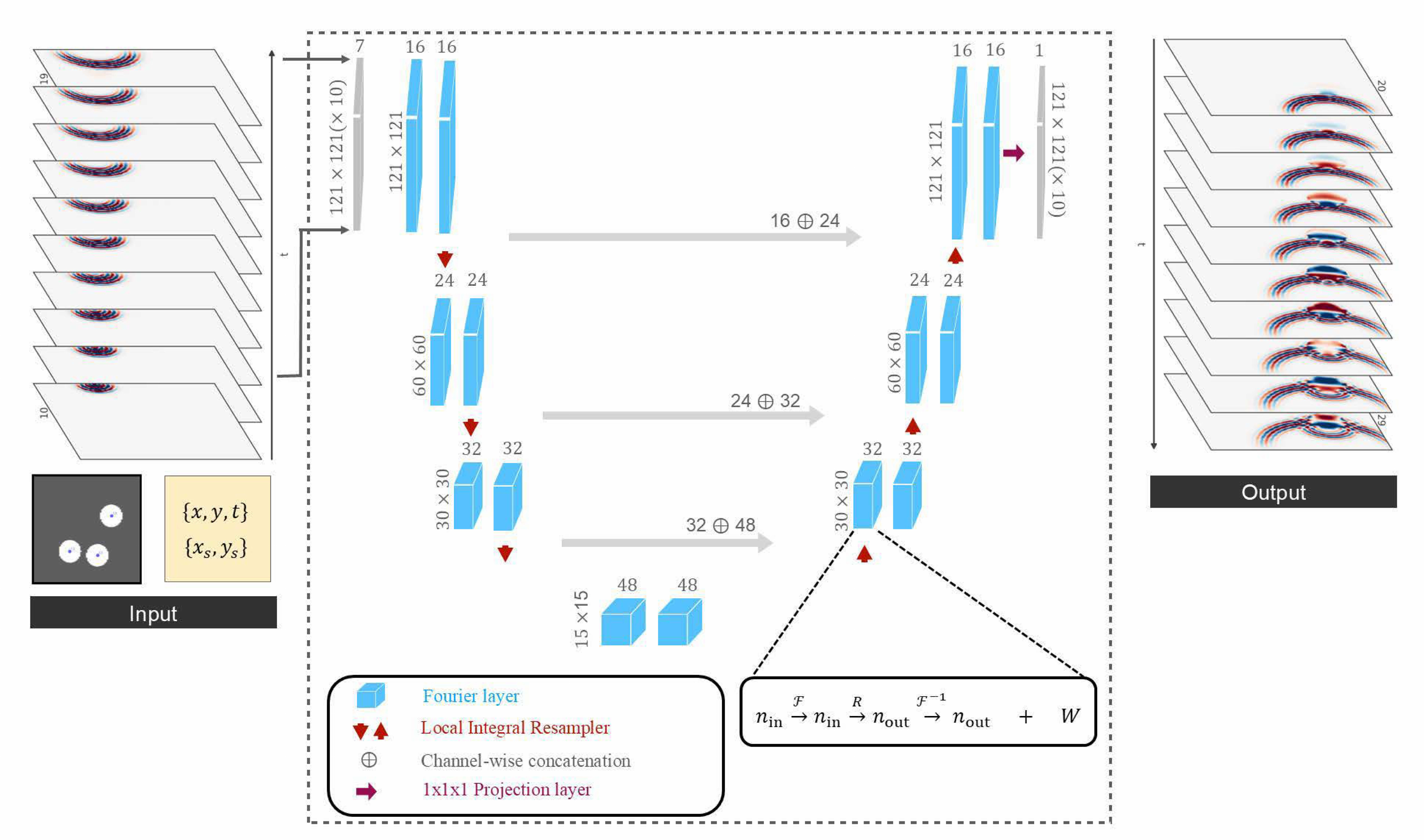}
    \caption{\small LU-NO: Time autoregressive neural surrogate for wavefields, using Fourier operator layers (blue) and local-integral resampling (red) to learn structurally coherent scale-invariant operator.}
    \label{fig:luno}
\end{figure}

\section{Problem Formulation}
\label{sec:problem_formulation}
Let $\Omega \subset \mathbb{R}^2$ denote a two-dimensional heterogeneous spatial domain of locations $\mathbf{x}=(x,y)\in\Omega$, with spatially varying wave velocity $c(\mathbf{x})$ characterizing the medium.
The task is to model the development of the ultrasonic pressure field $p(\mathbf{x},t)$ at location $\mathbf{x}$ and time $t$, 
given a historical temporal window and $c(\mathbf{x})$.
For more efficient learning, we additionally provide the model explicit spatial and temporal coordinates $(\mathbf{x},t)$ and information about the signal emission through the source coordinates $\mathbf{x}_s=(x_s,y_s)$ as extra input channels. This gives the surrogate model $\mathcal{G}$ as
\begin{align*}
\mathcal{G}:(p(\mathbf{x},t_{i:i+T-1}), c(\mathbf{x}),\mathbf{x},t,\mathbf{x}_s) \mapsto
p(\mathbf{x},t_{i+T:i+2T-1}),
\end{align*}
We map $T$ input frames directly to the next non-overlapping $T$ frames, rather than predicting a single future frame as in conventional autoregressive approaches~\cite{stender2023u}, to better capture longer-term temporal dynamics.
Fig~\ref{fig:luno} illustrates the setup, with the inputs on the left and the outputs on the right.

\section{Method}
We propose 
\emph{Local-Integral Resampler U-shaped Neural Operator (LU-NO)} (Fig.~\ref{fig:luno}) 
that learns an operator for the contiguous pressure fields,
using a combination of Fourier layers~\cite{li2021fourier, rahman2023uno} to capture global dependencies in the wavefield, a U-Net architecture~\cite{cciccek20163d, ronneberger2015u} to capture features across multiple spatial resolutions, and a new mechanism for transitioning between these resolutions using local-integral kernels~\cite{liu_schiaffini2024neural}. Next, we explain each component and its motivation.

\textbf{Fourier layer:} FNO~\cite{li2021fourier} is a neural operator architecture that parameterizes the non-local integral kernel of a neural operator layer~\cite{kovachki2023neural} in Fourier space. The key building block of FNO is the Fourier layer
\begin{align}
    v_{l+1}(\mathbf{x},t)= \sigma\left(\mathcal{F}^{-1}
\left(R_l \cdot \mathcal{F}(v_l)\right)(\mathbf{x},t)+W_l v_l(\mathbf{x},t)\right),
\label{eq:fourier}
\end{align}
where $v_l$ denotes the latent representation at layer $l$, $\mathcal{F}$ and $\mathcal{F}^{-1}$ are the Fourier transform and its inverse, $R_l$ contains the learnable weights applied to the retained Fourier modes, $W_l$ is a pointwise linear transformation, and $\sigma$ is a nonlinear activation. In our case, the Fourier transform is applied over both spatial dimensions and the temporal dimension. We use these layers within our LU-NO as the key components that operate within the U-Net resolution.

\textbf{U-Shaped architecture:} Pure FNO fails in heterogeneous media, as will be empirically shown in Section~\ref{sec:experiments}. Training the network to capture the pressure field characteristics at coarser resolution forces it also to model aspects that could be missed by the Fourier layers in isolation, such as reflections from secondary media. Consequently, we construct a U-shaped network architecture similar to the popular U-Net~\cite{cciccek20163d, ronneberger2015u}, but within each level we use the Fourier layers \eqref{eq:fourier}.
The architecture is similar to U-NO~\cite{rahman2023uno} that also uses Fourier layers within a U-Net-like architecture, but they used one Fourier layer at each level, requiring large skip connections. Instead, we use two Fourier layers at each level with skip connections between corresponding levels. We will later compare LU-NO against U-NO empirically in Section~\ref{sec:experiments}.

\textbf{Resampling:} Standard U-Net models use convolutions, pooling or interpolation for down- and upsampling across the layers. This holds also for U-NO, which uses
fixed coordinate mapping $(s_l: D_{l}\rightarrow D_{l+1})$, where $l$ is the layer index and $x\in D_l$ is mapped to the target location $s_l(x)\in D_{l+1}$, for spatial resizing, followed by standard interpolation.  
This is not optimal for direct modelling of the continuous system since the resampling depends on the input grid.

To address this, we use \emph{local-integral resampling} adapted from the local-integral kernel~\cite{liu_schiaffini2024neural}. Instead of interpolating or pooling latent features between grids, our re-sampler computes each target-grid value from a local neighborhood of the source grid using a learned kernel defined over spatial distance. The local kernel function at layer $l$ for $n^{\text{th}}$ channel is 
\begin{align*}
    k_{l,n}(d)= \sum_{m=1}^{M} \theta_{l,n,m}\phi_m(d),
\end{align*}
where $d$ is the distance between the source $D_{l}$ and target $D_{l+1}$ coordinates normalized to be in $[0,1]$, $\phi_m$ is the $m^{\text{th}}$ triangular basis function with learned coefficient
$\theta_{l,n,m}$. Given a feature function $v_{l,n}(s)$ on the source grid $s\in D_{l}$, the resampled feature $\tilde{v}_{l+1,n}(s_i')$, at $i^{\text{th}}$ target location $s_i'\in D_{l+1}$ is
\begin{align*}
    \tilde{v}_{l+1,n}(s_i') = \int_{s_i' - r}^{s_i'+r} k_{l,n}\left(|s_i' -s|\right)v_{l,n}(s)\,ds
\end{align*}
where only the source locations with $|s_i'-s|\leq r$ contribute to the local integral. Here $r\in (0,1]$ is the kernel radius, defined as a fixed fraction of the source-domain length, such that a fixed $r$ corresponds to the same fraction of the physical domain across different grid resolutions.
In practice, the integral is approximated by quadrature-based numerical integration. 

The operation is applied separably along the spatial dimensions, while preserving the temporal dimension. Intermediate spatial resolutions are fixed throughout the encoder-decoder hierarchy, whereas the final resampling layer dynamically maps the latent features back to the input spatial resolution.
When the kernel support extends beyond the domain boundary, only the source-grid points within the domain are included in the local integral (See App.~\ref{app:resampler_details} for details).

\section{Experiments}
\label{sec:experiments}
We compare our method against previous methods for 2D ultrasonic wave propagation, in an example of a water-filled domain containing three aluminum circles (see Fig.~\ref{fig:luno}). This simple domain allows illustrating qualitative differences between the methods already in modeling simple reflections. All methods are trained exclusively on limited low-resolution simulated data, consisting of only one domain structure with 14 distinct emission locations, but tested on an unseen structure and also on higher spatial resolution to evaluate generalization across domains and resolutions.

\begin{table}[!t]
\centering
\caption{\small Error on two spatial resolutions, with standard deviations over 16 test instances. Memory cost is for one training instance and the time is for full epoch, and \# Params indicates the number of trainable parameters.
}
\label{tab:model_results}

\resizebox{\columnwidth}{!}{%
\begin{tabular}{lccccc}
\hline
\multirow{2}{*}{Model}
& \multicolumn{2}{c}{Relative squared L2 error}
& \multirow{2}{*}{Mem. (GB)}
& \multirow{2}{*}{Time (s)}
& \multirow{2}{*}{\# Params.} \\
\cline{2-3}
& High resolution & Training resolution & & & \\
\hline

U-Net
& $1.160 \pm 0.025$
& $\mathbf{0.160 \pm 0.076}$
& 0.31
& 21
& 1.5M \\

FNO
& $0.405 \pm 0.073$
& $0.374 \pm 0.085$
& 17.32
& 59
& 944M \\

U-NO
& $0.453 \pm 0.047$
& $0.295 \pm 0.078$
& 16.66
& 61
& 1B \\

LU-NO
& $\mathbf{0.305 \pm 0.083}$
& $0.295 \pm 0.088$
& 11.91
& 34
& 739M \\

\hline
\end{tabular}%
}
\end{table}

\begin{figure*}[!t]
    \centering

    \begin{minipage}[c]{0.18\textwidth}
        \centering

        \includegraphics[width=\linewidth]
        {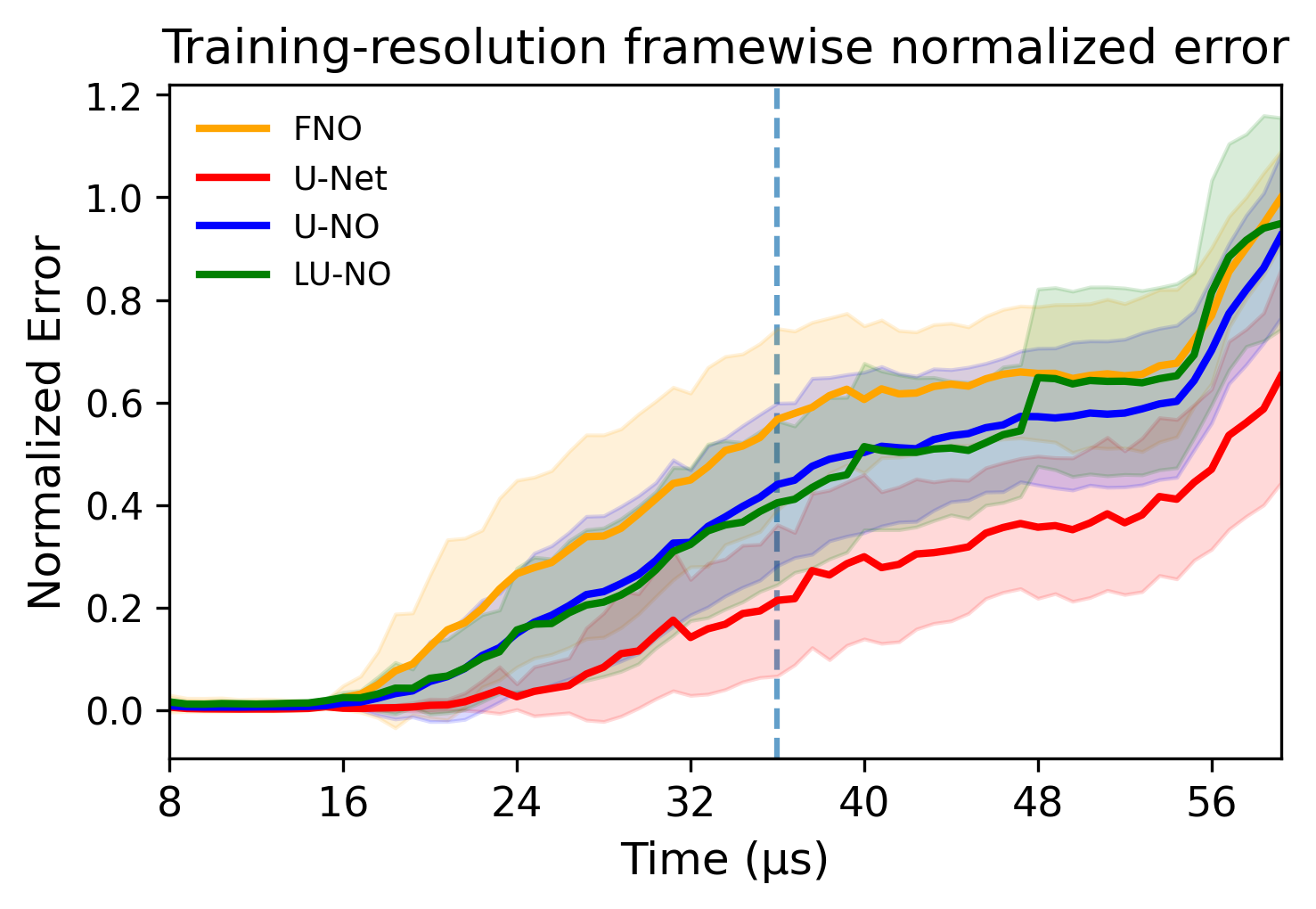}

        \textbf{(a)}

        \vspace{0.15cm}

        \includegraphics[width=\linewidth]
        {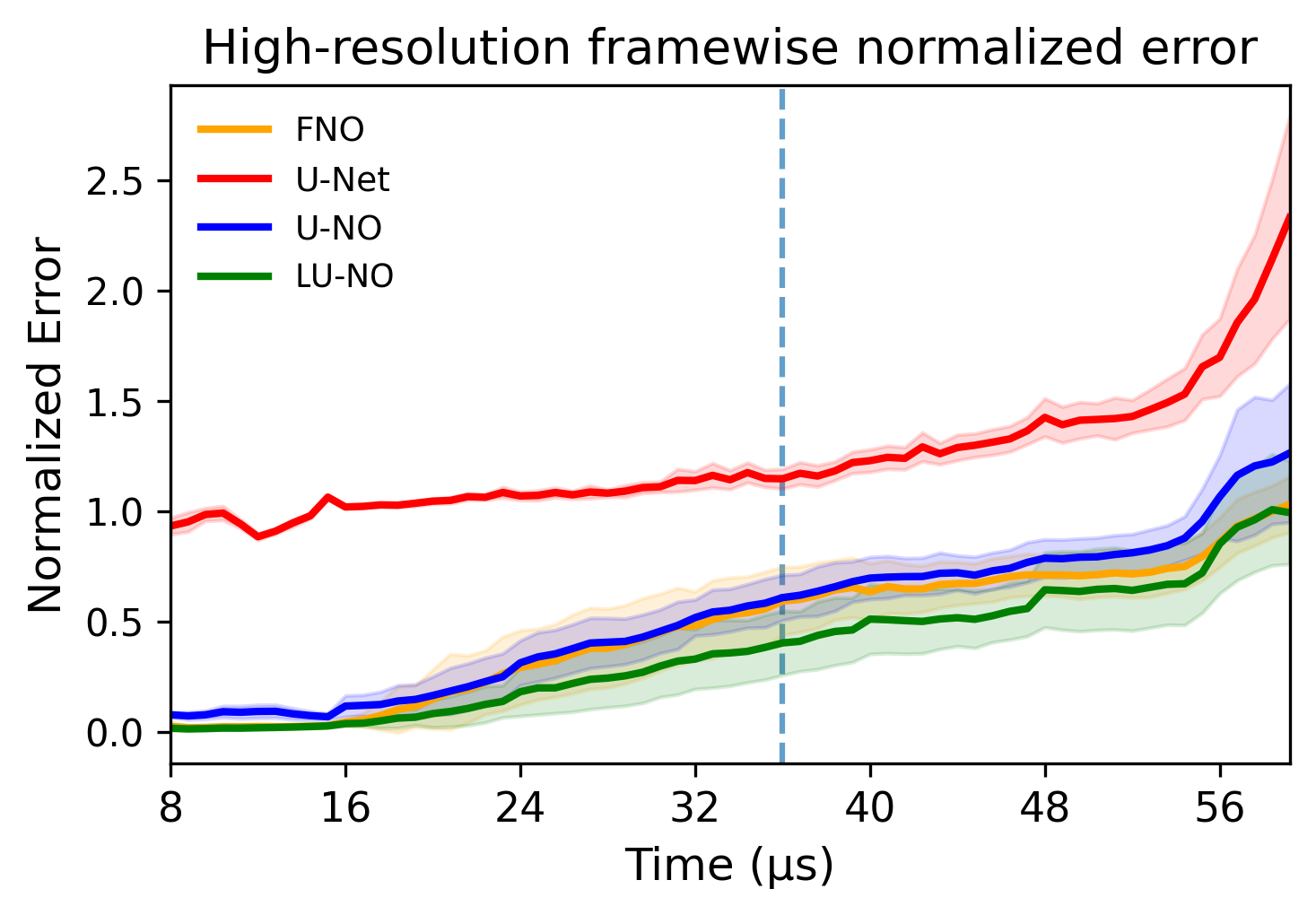}

        \textbf{(b)}
    \end{minipage}
    \hspace{0.015\textwidth}
    \begin{minipage}[c]{0.74\textwidth}
        \centering

        \includegraphics[width=\linewidth]
        {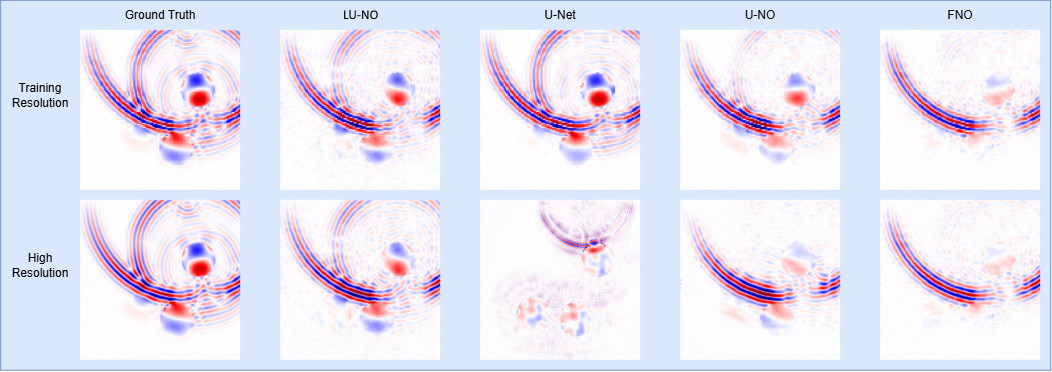}

        \textbf{(c)}
    \end{minipage}

    \caption{\small Frame-wise relative squared $L_2$ error over 16 test instances, computed as prediction SSE divided by the ground-truth signal energy at each frame. Solid lines show the mean, with shading indicating $\pm$ one standard deviation across test instances.
    \textbf{(a)} Training-resolution evaluation.
    \textbf{(b)} High-resolution evaluation.
    \textbf{(c)} Snapshots of $(t = 36\,\mu\mathrm{s})$ acoustic wave propagation. Our method best matches the ground truth in qualitative terms, as the only method that correctly models the main reflection also in high resolution (See App.~\ref{app:additional_results} for more results).}
    \label{fig:comparison}
\end{figure*}

\subsection{Data and evaluation setup}
\textbf{Simulation data:}
All simulations were done using COMSOL Multiphysics 6.3  
with the pressure-acoustics time-explicit interface and the elastic waves time-explicit interface.  
We simulate a rectangular water domain ($75\times75$mm) with three equally sized aluminum circles (16mm diameter), bounded by impedance boundary conditions to avoid reflections at the boundaries. An acoustic pressure source, located at the boundary of the domain, is actuated using a Hanning-windowed tone pulse featuring a 400 kHz center frequency. 
The simulated acoustic (and elastic) pressure wave fields were exported at a spatial resolution (discretization) of 0.25 mm, resulting in a domain of $301\times301$ samples. The simulated and exported temporal domain ranged from 0 to 60 $\mu\mathrm{s}$ with $0.2\,\mu\mathrm{s}$ resolution (interval).

\textbf{Training:} We use 14 simulation runs for training, all using the same structure (locations of the aluminum circles) but different emission locations, to demonstrate that the method can be trained on limited data. The training data was downsampled in space (x2.5) and time (x4) so that each simulation is a 75x121x121 tensor. To increase the variety, we perform simple data augmentation, with horizontal and vertical flips, to arrive in total of 42 augmented simulations, of which 38 were used for training and 4 left for testing in-domain generalization.
All models were trained using the setup in Section~\ref{sec:problem_formulation}
for 500 epochs using Adam with MSE, an initial learning rate of $10^{-3}$, halved every 50 epochs.

\textbf{Evaluation:}
To test out-of-domain generalization, we use 12 completely separate simulation runs, with different aluminum-circle locations, pooled together with the 4 instances extracted from the training data. We test the models both on the training resolution (121x121) and in the original simulation resolution (301x301) to evaluate spatial upscaling.
Performance is measured using frame-wise MSE and the relative squared $L_2$ error, $\|\hat{p}-p\|_2^2/\|p\|_2^2$, averaged over the predicted test simulations (See App.~\ref{app:metrics} for details).

\subsection{Baselines and architectural details}
We compare LU-NO against three neural surrogates representing three distinct approaches: Standard FNO~\cite{li2021fourier}, purely convolutional U-Net~\cite{li2025steady,stender2023u}, and U-NO~\cite{rahman2023uno} that uses Fourier layers within U-Net. For FNO and U-NO, we build on the NeuralOperator library~\cite{kossaifi2024library}. The details of all methods are summarized below and the full code is available at
\url{https://github.com/burlakorkmaz/Neural-Surrogates-for-ultrasound}.

\textbf{U-Net:}
Three encoder levels and a bottleneck with channel widths {\small $16\rightarrow32\rightarrow64\rightarrow128$}, followed by three decoder levels with widths $\small 64\rightarrow32\rightarrow16$. Each level contains two {\small $3\times3\times3$} convolutions with GroupNorm and GELU. Spatial downsampling uses {\small $1\times2\times2$} max pooling, while upsampling uses trilinear interpolation followed by skip-connection concatenation with the corresponding encoder features.

\textbf{FNO:}
Four Fourier layers with 64 hidden channels and 48 Fourier modes along each spatiotemporal dimension.

\textbf{U-NO:}
Fourteen Fourier layers arranged in an encoder-decoder hierarchy, with two consecutive layers at each channel width following {\small $16\rightarrow24\rightarrow32\rightarrow48\rightarrow32\rightarrow24\rightarrow16$} and 48 Fourier modes per dimension. The spatial dimensions are contracted and expanded by factors of two while preserving the temporal resolution, with skip connections between corresponding encoder and decoder levels.

\textbf{LU-NO:}
A U-shaped encoder-decoder uses channel widths within two consecutive layers  {\small $16\rightarrow24\rightarrow32\rightarrow48\rightarrow32\rightarrow24\rightarrow16$}, three skip connections, and Fourier layers with GroupNorm, GELU, and 48 modes per dimension. Spatial resampling follows {\small $121\rightarrow60\rightarrow30\rightarrow15\rightarrow30\rightarrow60\rightarrow121$} using learned local-integral resampling with 8 basis functions and radius $0.08$, while preserving the temporal dimension.

\subsection{Results}
Table~\ref{tab:model_results} summarizes the results numerically, but the qualitative differences between the models are best communicated by the visual results in Figure~\ref{fig:comparison}. The overall accuracy is the integral over the curves in sub-figures (a) and (b) showing how the accuracy develops over time, and the sub-figure (c) shows a snapshot of the predicted pressure fields at 36$\mu\mathrm{s}$ for inspection of which kind of mistakes the models do.

The main observation is that previous FNO-based methods are not sufficient for the task. Standard FNO does not capture the reflected or scattered waves at all even on the training resolution, and while U-NO is better on training resolution it fails in generalizing across spatial resolutions. In contrast, our LU-NO achieves correct spatial rescaling, verifying that it is the first method to achieve the promise of operator-based solutions for heterogeneous media. The difference between LU-NO and FNO-based methods is clear both quantitatively and qualitatively. As shown in Table~\ref{tab:model_results}, LU-NO is also more computationally efficient, both in terms of memory and time.

On the training resolution, U-Net is more accurate than LU-NO, best seen as more accurate reconstruction of longer wavelengths within the aluminum circles. Moreover, it is computationally efficient with substantially smaller model size. It is, however, solving an easier problem of learning the propagation for a given spatial resolution and high accuracy in this specific task is to be expected: Evaluation on a different spatial resolution reveals immediately that it fails to model the continuous system as such, with no sign of modeling the pressure field at all.

\begin{figure}[!t]
\centering

\begin{minipage}[c]{0.49\columnwidth}
    \centering
    \includegraphics[width=\linewidth]
    {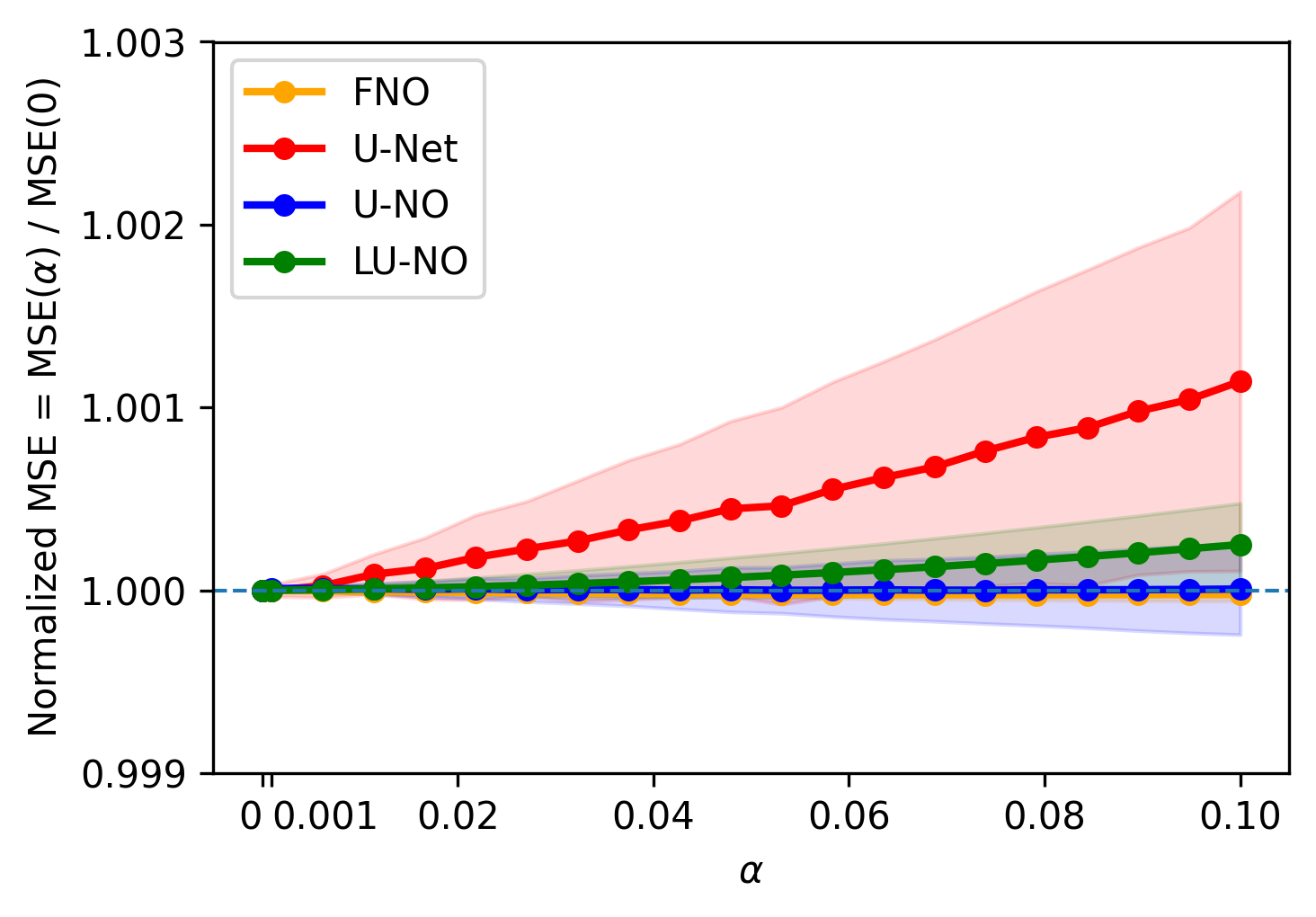}
\end{minipage}
\hfill
\begin{minipage}[c]{0.49\columnwidth}
    \centering
    \includegraphics[width=\linewidth]
    {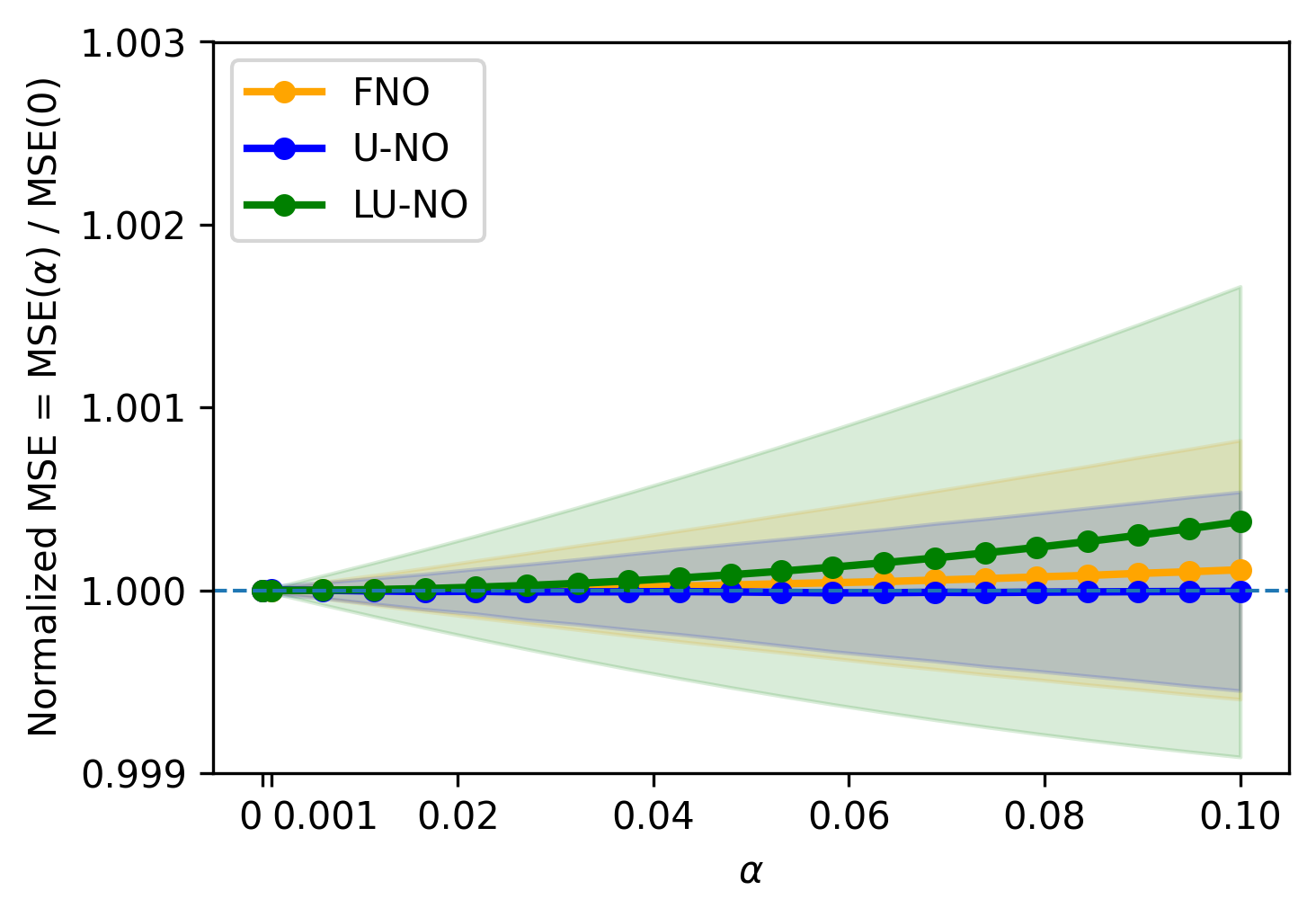}
\end{minipage}

\caption{\small Sensitivity to additive white Gaussian noise of $\sigma_n=\alpha\sigma_p$, where $\sigma_p$ is the RMSE of the mean-subtracted clean wavefield, for the initial pressure-field inputs (\textbf{Left:} Training resolution, \textbf{Right:} High resolution). The results are averaged over the 16 test simulations, with confidence interval (one std) over 10 independent noise seeds. U-Net with substantially worse overall accuracy is excluded from the high resolution plot.
}
\label{fig:noise_sensitivity}

\end{figure}

Figure~\ref{fig:noise_sensitivity} provides a complementary perspective for the simplications U-Net makes, by studying sensitivity for noisy initial pressure fields. The U-Net performance rapidly degrades when adding noise, whereas all FNO-based methods that model the continuous physical problem remains robust to noise. 
This robustness may be attributed to the global Fourier layers that aggregate information across the entire domain rather than relying only on local neighborhoods captured by the convolutions.

\section{Conclusion}
We introduced LU-NO, a U-shaped neural operator combining Fourier layers with learned local-integral resampling for ultrasonic wave propagation across spatial resolutions. We showed that, in contrast to previous FNO-based methods, it can correctly model reflections and refractions in heterogeneous medium and demonstrated the evaluation accuracy remains intact when changing the resolution, suggesting it learns the underlying continuous system correctly as an operator.
The accuracy does not yet reach that of a simpler convolutional U-Net, that only learns the pressure field at a given spatial resolution, but it is worth noting that in this work we intentionally trained the models on very limited data that may not be sufficient for learning the larger models perfectly.

\section{Acknowledgement}
BNK, HVT and AK acknowledge the research environment provided by ELLIS Institute Finland. The work received funding from the Jane and Aatos Erkko Foundation, through the Transcranial AI-Powered Ultrasound Computed Tomography grant.

\vfill\pagebreak

\small
\bibliographystyle{IEEEbib}
\bibliography{refs}

@inproceedings{li2021fourier,
  author    = {Z. Li and N. Kovachki and K. Azizzadenesheli and B. Liu and K. Bhattacharya and A. Stuart and A. Anandkumar},
  title     = {{Fourier Neural Operator for Parametric Partial Differential Equations}},
  booktitle = {Proc. of the International Conference on Learning Representations},
  year      = {2021}
}

@article{kovachki2023neural,
  author  = {N. Kovachki and Z. Li and B. Liu and K. Azizzadenesheli and K. Bhattacharya and A. Stuart and A. Anandkumar},
  title   = {Neural {O}perator: {L}earning Maps Between Function Spaces with Applications to {PDEs}},
  journal = {Journal of Machine Learning Research},
  volume  = {24},
  number  = {89},
  pages   = {1--97},
  year    = {2023}
}

@inproceedings{cciccek20163d,
  author    = {{\"O}. {\c{C}}i{\c{c}}ek and A. Abdulkadir and S. S. Lienkamp  and T. Brox and O. Ronneberger},
  title     = {{3D U-Net:} {L}earning Dense Volumetric Segmentation from Sparse Annotation},
  booktitle = {Proc. of the International Conference on Medical Image Computing and Computer-Assisted Intervention},
  year      = {2016}
}

@inproceedings{ronneberger2015u,
  author    = {O. Ronneberger and P. Fischer and T. Brox},
  title     = {{U-Net: C}onvolutional Networks for Biomedical Image Segmentation},
  booktitle = {Proc. of the International Conference on Medical Image Computing and Computer-Assisted Intervention},
  year      = {2015}
}

@article{rahman2023uno,
  author  = {M. A. Rahman and Z. E. Ross and K. Azizzadenesheli},
  title   = {{U-NO: U}-Shaped Neural Operators},
  journal = {Transactions on Machine Learning Research},
  year    = {2023}
}

@inproceedings{liu_schiaffini2024neural,
  author    = {M. Liu-Schiaffini and J. Berner and B. Bonev and T. Kurth and K. Azizzadenesheli and A. Anandkumar},
  title     = {Neural Operators with Localized Integral and Differential Kernels},
  booktitle = {Proc. of the 41st International Conference on Machine Learning},
  year      = {2024}
}

@article{stender2023u,
  author  = {M. Stender and J. Ohlsen and H. Geisler and A. Chabchoub and N. Hoffmann and A. Schlaefer},
  title   = {{UP-Net}: A Generic Deep Learning-Based Time Stepper for Parameterized Spatio-Temporal Dynamics},
  journal = {Computational Mechanics},
  volume  = {71},
  number  = {6},
  pages   = {1227--1249},
  year    = {2023}
}

@article{li2025steady,
  author  = {X. Li and L. Liu and J. Lu and Y. Zhang and D. Cong and Y. Xiang},
  title   = {A Steady-State Wavefield Polarity-Aware {U-Net} for Ultrasonic Guided Waves Depth-Resolved Imaging},
  journal = {NDT \& E International},
  volume  = {158},
  pages   = {103597},
  year    = {2026}
}

@article{li2023solving,
  author  = {B. Li and H. Wang and S. Feng and X. Yang and Y. Lin},
  title   = {Solving Seismic Wave Equations on Variable Velocity Models with {F}ourier Neural Operator},
  journal = {IEEE Transactions on Geoscience and Remote Sensing},
  volume  = {61},
  pages   = {1--18},
  year    = {2023}
}

@article{huang2025learned,
  author  = {X. Huang and T. Alkhalifah},
  title   = {Learned Frequency-Domain Scattered Wavefield Solutions Using Neural Operators},
  journal = {Geophysical Journal International},
  volume  = {241},
  number  = {3},
  pages   = {1467--1478},
  year    = {2025}
}

@article{zhang2023learning,
  author  = {T. Zhang and D. Trad and K. Innanen},
  title   = {Learning to Solve the Elastic Wave Equation with {F}ourier Neural Operators},
  journal = {Geophysics},
  volume  = {88},
  number  = {3},
  pages   = {T101--T119},
  year    = {2023}
}

@inproceedings{raonic2023convolutional,
  author    = {B. Raoni{\'c} and R. Molinaro and T. De Ryck and T. Rohner and F. Bartolucci and R. Alaifari and S. Mishra and E. de B{\'e}zenac},
  title     = {Convolutional Neural Operators for Robust and Accurate Learning of {PDEs}},
  booktitle = {Advances in Neural Information Processing Systems},
  year      = {2023}
}

@article{kossaifi2024library,
  author  = {J. Kossaifi and N. Kovachki and Z. Li and D. Pitt and M. Liu-Schiaffini and R. J. George and B. Bonev and K. Azizzadenesheli and J. Berner and V. Duruisseaux and A. Anandkumar},
  title   = {A Library for Learning Neural Operators},
  journal = {Journal of Machine Learning Research},
  volume  = {27},
  number  = {205},
  pages   = {1--6},
  year    = {2026}
}

@article{rose2004ultrasonic,
  author  = {J. L. Rose},
  title   = {Ultrasonic Guided Waves in Structural Health Monitoring},
  journal = {Key Engineering Materials},
  volume  = {270--273},
  number  = {I},
  pages   = {14--21},
  year    = {2004}
}

@article{korkmaz2026physical,
  author  = {B. N. Korkmaz and S. Gharib and H. Wei and J. Korsimaa and P. Salminen and M. Weber and D. Iablonskyi and A. Salmi and A. Klami},
  title   = {Physical Simulator-Based Neural Networks for Real-Time Fouling Tomography},
  journal = {ACM Transactions on Probabilistic Machine Learning},
  volume  = {2},
  number  = {3},
  pages   = {1--25},
  year    = {2026}
}

@article{gharib2025self,
  author  = {S. Gharib and D. Iablonskyi and J. Mustonen and J. Korsimaa and P. Salminen and B. N. Korkmaz and M. Weber and A. Salmi and A. Klami},
  title   = {Self-Adaptation of Ultrasound Sensing Networks},
  journal = {Mechanical Systems and Signal Processing},
  volume  = {224},
  pages   = {112214},
  year    = {2025}
}

@article{xiao2025beamforming,
  author  = {D. Xiao and A. C. H. Yu},
  title   = {Beamforming-Integrated Neural Networks for Ultrasound Imaging},
  journal = {Ultrasonics},
  volume  = {145},
  pages   = {107474},
  year    = {2025}
}

@article{yan2025untrained,
  author  = {W. Yan and Q. Zhang and Y. Wu and Z. Liu and H. Zhang and Z. Wang and J. Yuan and M. Ding and M. Yuchi and W. Qiu},
  title   = {Untrained Neural Network-Based Full-Waveform Inversion for Breast Sound Speed Imaging in Ultrasound Computed Tomography},
  journal = {IEEE Transactions on Instrumentation and Measurement},
  volume  = {74},
  pages   = {1--14},
  year    = {2025}
}

@inproceedings{khan2022unfolding,
  author    = {C. Khan and R. J. G. van Sloun and B. Byram},
  title     = {Unfolding Model-Based Beamforming for High Quality Ultrasound Imaging},
  booktitle = {Proc. of the 2022 IEEE International Conference on Acoustics, Speech and Signal Processing},
  year      = {2022}
}

@inproceedings{ruan2024delineation,
  author    = {Y. Ruan and R. Jin and Z. Liu and C. Wang and L. Zhang and T. Peng},
  title     = {Delineation of Prostate Cancer via Enhanced {AI}-Based Algorithm in Ultrasound Images},
  booktitle = {Proc. of the 2024 IEEE International Conference on Acoustics, Speech and Signal Processing},
  year      = {2024}
}

@article{moilanen2024software,
  author  = {P. Moilanen and T. Rauhala and S. Ahmadzai},
  title   = {Software-Guided Clamp-On Power Ultrasound Solution for Fouling Mitigation in Tubular Heat Exchangers},
  journal = {Heat and Mass Transfer},
  volume  = {60},
  number  = {5},
  pages   = {769--776},
  year    = {2024}
}

@inproceedings{sharma2022deep,
  author    = {J. K. Sharma and R. Soman and P. Kudela and E. Chatzi and W. Ostachowicz},
  title     = {A Deep Learning Based Surrogate Modelling for Wave Propagation in Structures},
  booktitle = {Proc. of Sensors and Smart Structures Technologies for Civil, Mechanical, and Aerospace Systems},
  year      = {2022}
}

@article{augenstein2023neural,
  author  = {Y. Augenstein and T. Rep{\"a}n and C. Rockstuhl},
  title   = {Neural Operator-Based Surrogate Solver for Free-Form Electromagnetic Inverse Design},
  journal = {ACS Photonics},
  volume  = {10},
  number  = {5},
  pages   = {1547--1557},
  year    = {2023}
}

@article{gottlieb1997gibbs,
  author  = {D. Gottlieb and C.W. Shu},
  title   = {On the {G}ibbs Phenomenon and Its Resolution},
  journal = {SIAM Review},
  volume  = {39},
  number  = {4},
  pages   = {644--668},
  year    = {1997}
}

@article{davis2026fourier,
  author  = {O. Davis and M. Motamed and O. Runborg},
  title   = {Fourier Residual Networks Achieve Spectral Accuracy for Discontinuous Functions},
  journal = {arXiv:2605.03549},
  year    = {2026}
}

@article{guasch2020full,
  author  = {L. Guasch and O. Calder{\'o}n Agudo and M. X. Tang and P. Nachev and M. Warner},
  title   = {Full-Waveform Inversion Imaging of the Human Brain},
  journal = {npj Digital Medicine},
  volume  = {3},
  number  = {1},
  pages   = {28},
  year    = {2020}
}

@article{zhang2024cross,
  author  = {N. Zhang and Y. Zhao and Y. Yuan and Y. Xiao and M. Qin and Y. Shen},
  title   = {Cross-Correlation Adjustment Full-Waveform Inversion with Source Encoding in Ultrasound Computed Tomography},
  journal = {Ultrasonics},
  volume  = {142},
  pages   = {107392},
  year    = {2024}
}

\onecolumn
\appendix

\section{Domain structure}
\label{app:domain}
\begin{figure*}[!h]
\centering

\begin{minipage}[c]{0.23\textwidth}
    \centering
    \includegraphics[width=\linewidth]
    {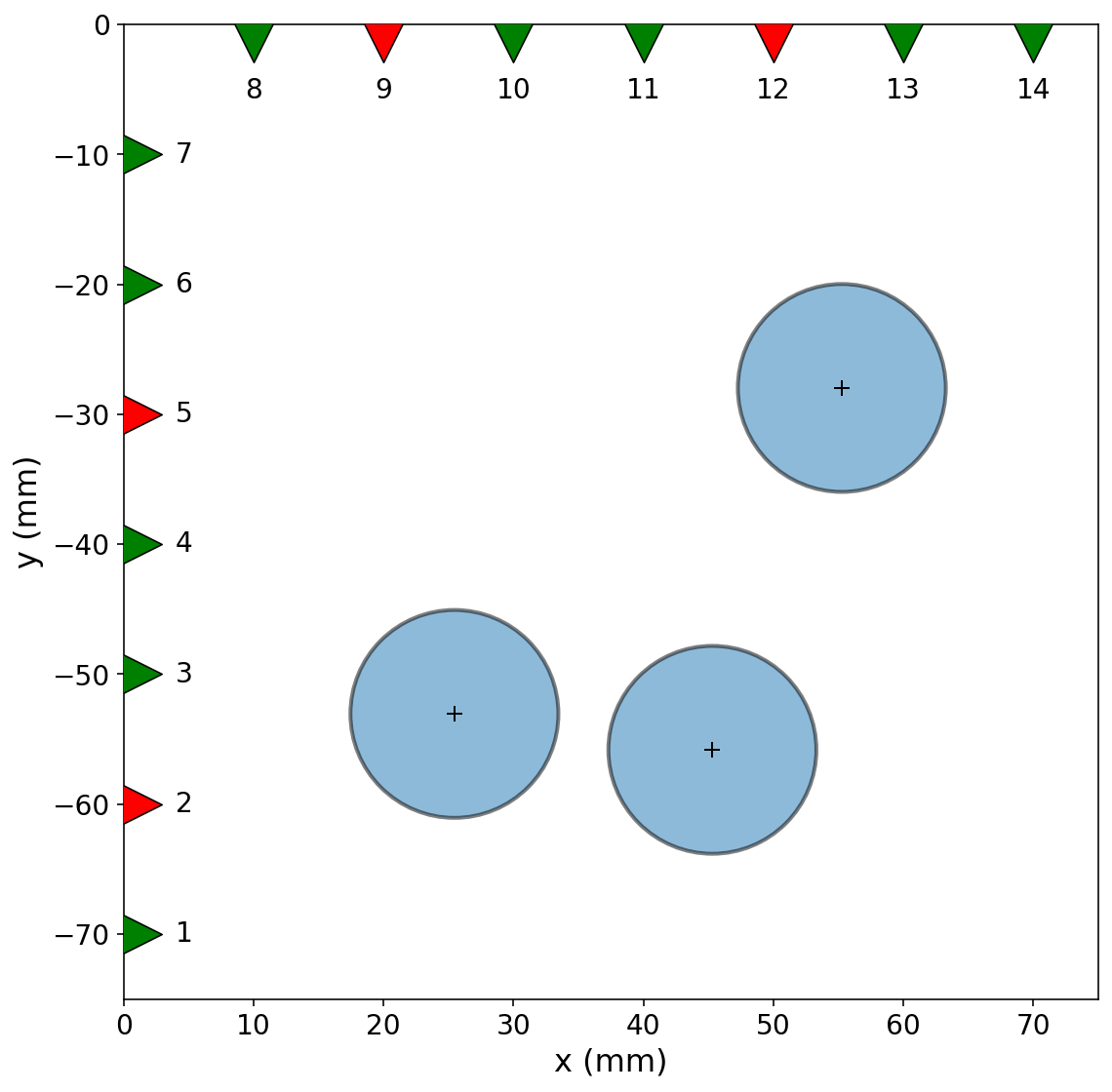}
\end{minipage}
\hspace{0.01\textwidth}
\begin{minipage}[c]{0.23\textwidth}
    \centering
    \includegraphics[width=\linewidth]
    {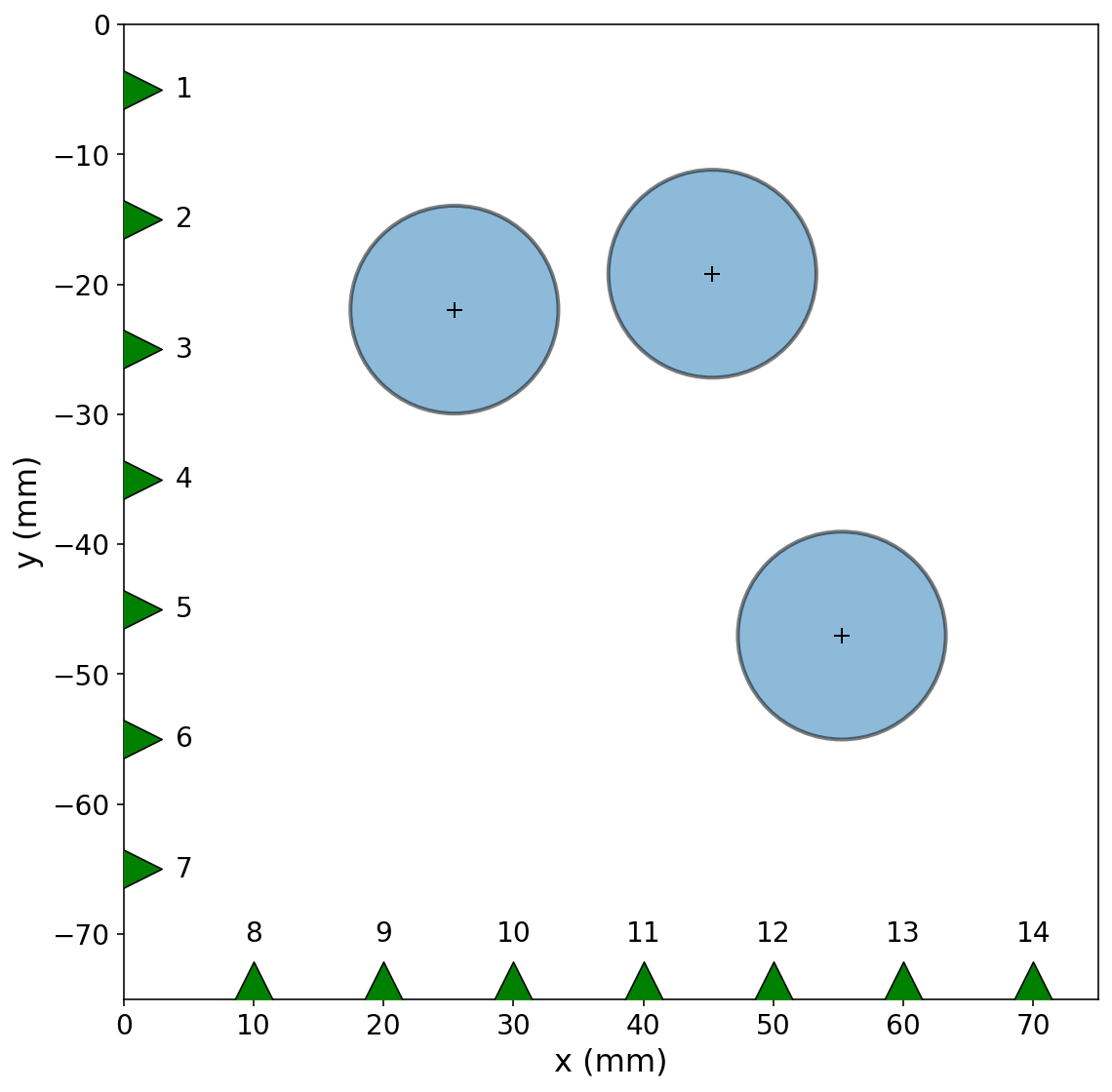}
\end{minipage}
\hspace{0.01\textwidth}
\begin{minipage}[c]{0.23\textwidth}
    \centering
    \includegraphics[width=\linewidth]
    {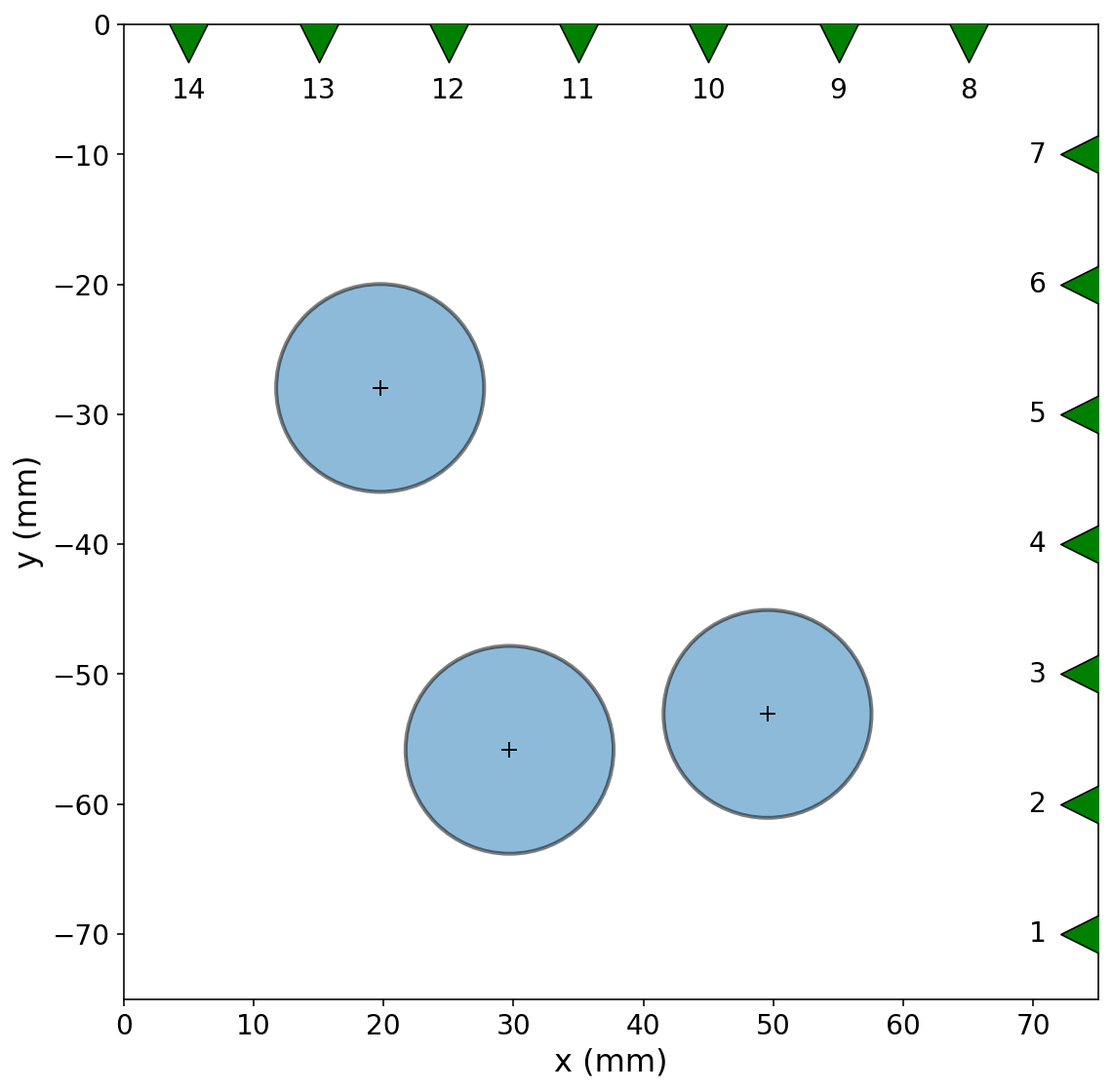}
\end{minipage}
\hspace{0.01\textwidth}
\begin{minipage}[c]{0.23\textwidth}
    \centering
    \includegraphics[width=\linewidth]
    {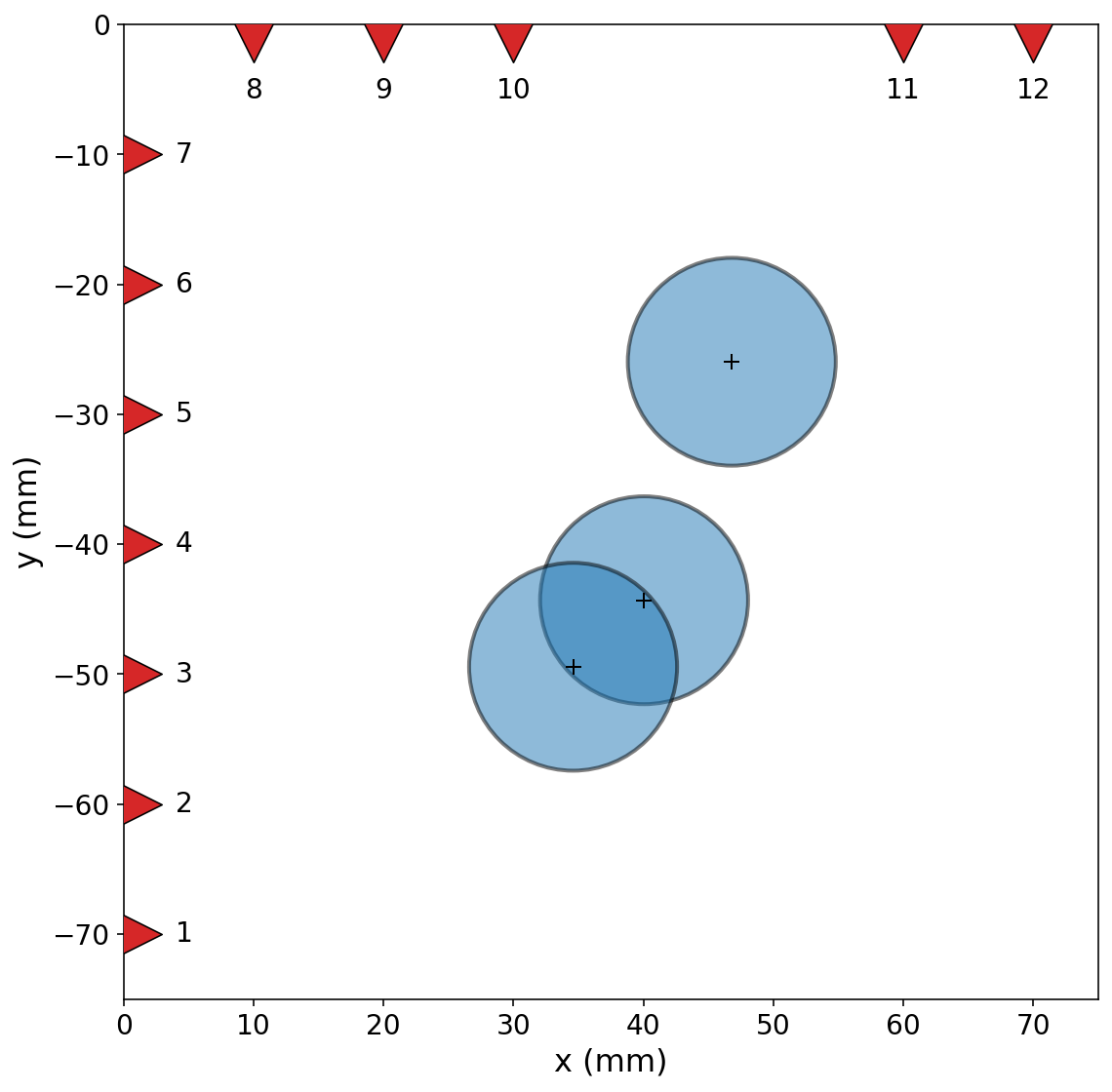}
\end{minipage}

\caption{\small Illustration of the physical domain.
(\textbf{First three plots:} In-domain structure,
\textbf{Last plot:} Out-of-domain structure.)
Arrows indicate pulse locations; green ones are used during training,
and red ones are used for testing. Blue circles show aluminum areas.}
\label{fig:physical_domain}

\end{figure*}

Figure~\ref{fig:physical_domain} illustrates the physical domains used in the experiments. The first three illustrations show the in-domain geometry used for training, consisting of the original domain and its vertically and horizontally flipped versions. The last panel shows the out-of-domain geometry used for testing. The circles represent the aluminum area, while the arrows indicate the pulse locations, with green and red arrows denoting training samples and testing samples, respectively.

\section{Additional results}
\label{app:additional_results}
\begin{figure*}[!h]
    \centering

    \includegraphics[width=0.95\textwidth]
    {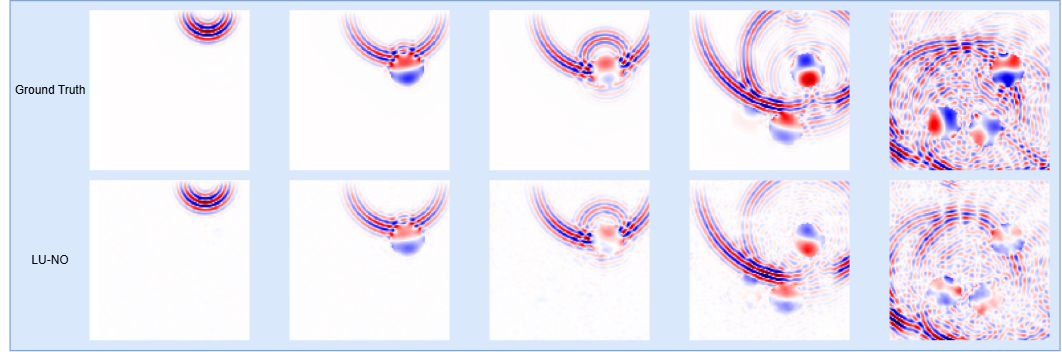}

    \caption{\small Snapshots of acoustic wave propagation at
    different time points.}
    \label{fig:comparison2}
\end{figure*}

Figure~\ref{fig:comparison2} illustrates the evaluation of LU-NO at different time points. For additional full time-domain predictions, see \url{https://github.com/burlakorkmaz/Neural-Surrogates-for-ultrasound}. 

\section{Local-integral resampling details}
\label{app:resampler_details}

The local-integral resampler operates independently on each feature
channel and is applied separably along the two spatial dimensions.
Let $a\in\{x,y\}$ denote a spatial dimension and let
$s^{(a)}\in[0,1]$ denote its normalized coordinate. For a source grid
containing $N_a$ points, the normalized source coordinates are
defined as
\begin{align*}
s_j^{(a)}
=
\frac{j}{N_a-1},
\qquad
j=0,\ldots,N_a-1.
\end{align*}
Thus, the kernel radius $r$ represents a fixed fraction of the spatial domain and is independent of the number of grid points. Similarly, for a target grid containing $N_a'$ points, the
normalized target coordinates are defined as
\begin{align*}
s_i^{\prime(a)}
=
\frac{i}{N_a'-1},
\qquad
i=0,\ldots,N_a'-1.
\end{align*}
The continuous kernel for feature channel $n$ at layer $l$ is
represented using $M$ triangular basis functions,
\begin{align*}
k_{l,n}(d)
=
\sum_{m=1}^{M}
\theta_{l,n,m}\phi_m(d),
\end{align*}
where the basis centers and width are $\mu_m = \frac{(m-1)r}{M-1}$, $\delta = \frac{r}{M-1}$ and $m=1,\ldots,M$. Then the triangular basis functions are defined as
\begin{align*}
\phi_m(d)
=
\max\left(
0,\,
1-\frac{|d-\mu_m|}{\delta}
\right).
\end{align*}
For either spatial dimension $a\in\{x,y\}$, the continuous local-integral operation is
\begin{align*}
\tilde{v}_{l+1,n}(s_i^{\prime(a)})
=
\int_{
D\cap
[s_i^{\prime(a)}-r,\,
 s_i^{\prime(a)}+r]
}
k_{l,n}
\left(
\left|s_i^{\prime(a)}-s^{(a)}\right|
\right)
v_{l,n}(s^{(a)})\,ds^{(a)},
\end{align*}
where $D=[0,1]$. The intersection with $D$ truncates the kernel
support at the domain boundaries, such that no values outside the
domain contribute to the integral. For a uniformly spaced source grid, the integral is approximated
by numerical quadrature as
\begin{align*}
\tilde{v}_{l+1,n}(s_i^{\prime(a)})
\approx
\Delta s^{(a)}
\sum_{j\in\mathcal{F}_r(s_i^{\prime(a)})}
k_{l,n}\left(
\left|s_i^{\prime(a)}-s_j^{(a)}\right|
\right)
v_{l,n}(s_j^{(a)}),
\end{align*}
\begin{align*}
\mathcal{F}_r(s_i^{\prime(a)})
=
\left\{
j:
\left|s_i^{\prime(a)}-s_j^{(a)}\right|\le r,\;
s_j^{(a)}\in D
\right\}.
\end{align*}
Here, $\Delta s^{(a)}
=
\frac{1}{N_a-1}$ denotes the spacing between
consecutive points of the uniformly spaced normalized source grid
and serves as the quadrature weight. The indices $i$ and $j$ correspond to
the target-grid and source-grid points, respectively, and $\mathcal{F}_r$ determines which source points to participate. Boundary handling is therefore performed
directly by restricting the summation to valid source-grid points
within both the domain and the local kernel support.
Finally, the two-dimensional resampling is applied separably,
\begin{align*}
\mathcal{R}_{\mathrm{2D}}
=
\mathcal{R}_{y}\circ\mathcal{R}_{x},
\end{align*}
where the same channel-wise learned kernel parameters are used along
both spatial dimensions. The resampling changes only the spatial
discretization, while the number of feature channels and temporal
resolution remain unchanged.

\section{Metrics}
\label{app:metrics}
The relative squared $L_2$ error $E_{\mathrm{rel}}$ reported in Table~\ref{tab:model_results} computed for a predicted simulation as below
\begin{align*}
E_{\mathrm{rel}}
=
\frac{\|\hat{p}-p\|_2^2}{\|p\|_2^2},
\end{align*}

\begin{align*}
E_{\mathrm{rel}}
=
\frac{
\sum_{t=1}^{N_t}\sum_{\mathbf{x}\in\Omega_h}
\left(\hat{p}(\mathbf{x},t)-p(\mathbf{x},t)\right)^2
}{
\sum_{t=1}^{N_t}\sum_{\mathbf{x}\in\Omega_h}
p(\mathbf{x},t)^2
}.
\end{align*}
MSE is computed for a predicted simulation to be used in Figure~\ref{fig:noise_sensitivity} as

\begin{align*}
\mathrm{MSE}
=
\frac{1}{N_t|\Omega_h|}
\sum_{t=1}^{N_t}
\sum_{\mathbf{x}\in\Omega_h}
\left(
\hat{p}(\mathbf{x},t)-p(\mathbf{x},t)
\right)^2,
\end{align*}
where $N_t$ is the total number of evaluated frames and $\Omega_h$ is the discrete spatial grid. Then for sensitivity analyses, we normalized MSE as
\begin{align*}
\mathrm{MSE}_{\mathrm{norm}}(\alpha)
=
\frac{\mathrm{MSE}(\alpha)}
{\mathrm{MSE}(0)},
\end{align*}
where $\mathrm{MSE}(\alpha)$ denotes the MSE at noise level $\alpha$, and
$\mathrm{MSE}(0)$ denotes the corresponding MSE without added noise.

\end{document}